\documentclass[10pt]{article}

\usepackage[a4paper,margin=25mm]{geometry}
\usepackage{amsmath,amssymb,amsthm}
\usepackage{bm}
\usepackage{graphicx}
\usepackage{xcolor}

\newtheorem{lemma}{Lemma}
\newtheorem{proposition}[lemma]{Proposition}
\newtheorem{theorem}[lemma]{Theorem}
\newtheorem{corollary}[lemma]{Corollary}

\newcommand{\R}{\mathbb{R}}
\newcommand{\E}{\mathbb{E}}
\newcommand{\tr}{\operatorname{tr}}
\newcommand{\vct}{\operatorname{vec}}
\newcommand{\Sym}{\operatorname{Sym}}
\newcommand{\Skew}{\operatorname{Skew}}

\title{A Frame-Theoretic Criterion for Three-Factor Learning Rules,\\
with Oja's Subspace Rule as a Worked Case}
\author{Taiki Yamada}
\date{\today}

\begin{document}
\maketitle

\begin{abstract}
We give a sufficient condition under which a learning rule admits a three-factor
implementation whose global factor is available in closed form.
Let $\xi$ be a frame for a subspace $V$ with frame operator $S$.
If the flow of a learning rule can be written as $S$ applied to a vector $m \in V$,
then the coefficient of its frame expansion is simply $(m,\xi)$:
the frame operator cancels, and the global factor is an inner product with $m$.
We show that Oja's subspace rule satisfies this condition.
Taking $\xi = \vct(xx^\top) - \vct(\Sigma)$, which is a frame for the space of symmetric
matrices under Gaussian inputs, right-preconditioning by $\Sigma$ maps Oja's flow exactly
onto $S\,\vct\big(\tfrac{1}{2}(I - W^\top W)\big)$; the preconditioning is not a device but
the operation that realizes the condition.
The resulting global factor is
$g(x,u) = \big(\tfrac{1}{2}(I - W^\top W),\, xx^\top - \Sigma\big)_F$,
which is precisely the third factor of the error-gated Hebbian rule for PCA (EGHR-PCA).
The known equivalence between EGHR-PCA and Oja's subspace rule follows as a corollary,
and the structural property responsible for it becomes explicit.
\end{abstract}

\section{Introduction}

Oja's subspace rule \cite{oja1982simplified,oja1989neural} is a classical learning rule for
principal subspace extraction in linear neural networks.
While mathematically elegant, it requires synapse-specific feedback that is difficult to
implement in biological neural circuits.

Isomura and Toyoizumi \cite{isomura2018error} proposed EGHR-PCA, a three-factor learning
rule \cite{kusmierz2017learning,fremaux2016neuromodulated} that replaces this local feedback
with a global scalar modulator multiplying the local Hebbian term.
They proved that the two rules are equivalent under Gaussian inputs.
However, a principled derivation of the global factor was not given; its construction
appears to be heuristic.
The question of how to obtain biologically implementable rules from mathematically canonical
ones in a systematic way has also been approached from the side of objective functions, most
notably through similarity matching \cite{pehlevan2015hebbian}, where local Hebbian and
anti-Hebbian rules are derived by minimizing a principled cost.

In this note we approach the question from a different direction and isolate a structural
condition rather than an objective.
Frame theory \cite{christensen2008frames,ali1993continuous} allows a vector to be expanded
against an overcomplete family; the expansion coefficient is exactly what a three-factor rule
needs as its global factor.
Our observation is that this coefficient becomes elementary precisely when the flow of the
learning rule is the image of a natural vector under the frame operator
(Lemma~\ref{lem:cancel}).
Oja's subspace rule satisfies this condition once it is right-preconditioned by the input
covariance, and the resulting coefficient is the third factor of EGHR-PCA
(Theorem~\ref{thm:main}).
The equivalence established in \cite{isomura2018error} is recovered as a corollary, and the
reason behind it---namely, that the preconditioned Oja flow lies in the image of the frame
operator---becomes visible.

Since the condition is stated for a general frame and a general flow, the argument does not
require prior knowledge of the target learning rule.
Section~\ref{sec:scope} discusses where its hypotheses fail.

\section{Setup}
\label{sec:pre}

Consider a single-layer linear neural network.
Let $x \in \R^{N_x}$ be the input vector, assumed Gaussian with zero mean and covariance
matrix $\Sigma \in \R^{N_x \times N_x}$, and let $u = Wx \in \R^{N_u}$ be the output vector,
where $W \in \R^{N_u \times N_x}$ is the synaptic weight matrix to be trained.
We write $\E_x[\cdot]$ for the expectation over the distribution of $x$.

We use the column-stacking vectorization $\vct$, so that
$\vct(X)_{(j-1)N_x + i} = X_{ij}$; equivalently, $\vct(ab^\top) = b \otimes a$ for
$a,b \in \R^{N_x}$, where $\otimes$ denotes the Kronecker product.
In particular, $\vct(xx^\top) = x \otimes x$.
Throughout we use the identity $\vct(AXB) = (B^\top \otimes A)\vct(X)$
\cite{magnus2019matrix}.
The standard inner product on $\R^{N_x^2}$ is written $(\cdot,\cdot)$, and the Frobenius
inner product on matrices is $(A,B)_F = \tr(A^\top B)$; these agree under $\vct$, that is,
$(\vct A, \vct B) = (A,B)_F$.
We write $\Sym$ and $\Skew$ for the spaces of symmetric and skew-symmetric
$N_x \times N_x$ matrices, respectively.

\subsection{Oja's subspace rule}
Oja's subspace rule \cite{oja1989neural} is the learning rule for $W$ defined by
\begin{alignat}{1}
    \label{Eq:Oja}
    \dot{W}^{(\text{Oja})} &= \E_x\big[u(x - W^\top u)^\top\big],\\
    \label{Eq:Oja-ele}
    \dot{W}_{ij}^{(\text{Oja})} &= \E_x\big[u_i(x_j - (W^\top u)_j)\big].
\end{alignat}
Carrying out the expectation gives the closed form
\begin{alignat}{1}
    \label{Eq:Oja-closed}
    \dot{W}^{(\text{Oja})} = W\Sigma(I - W^\top W).
\end{alignat}
Implementing Eq.~\eqref{Eq:Oja-ele} in a neural circuit requires synapse-specific feedback,
because the term $(W^\top u)_j$ depends on the whole weight matrix rather than on the
pre- and postsynaptic activity of synapse $ij$ alone.

\subsection{The error-gated Hebbian rule for PCA}
EGHR-PCA, proposed by Isomura and Toyoizumi \cite{isomura2018error}, removes this
requirement.
It is defined by
\begin{alignat}{1}
    \label{Eq:Isomura}
    \dot{W}^{(\text{EGHR})} &= \E_x\big[g(x,u)\, ux^\top\big],\\
    \label{Eq:Isomura-ele}
    \dot{W}_{ij}^{(\text{EGHR})} &= \E_x\big[g(x,u)\, u_i x_j\big],
\end{alignat}
where the global scalar factor $g(x,u) \in \R$ is
\begin{alignat}{1}
    \label{Eq:Isomura-global}
    g(x,u) = \frac{1}{2}\left( |x|^2 - |u|^2 - \E_x\big[|x|^2 - |u|^2\big] \right).
\end{alignat}
Eq.~\eqref{Eq:Isomura-ele} is a three-factor learning rule
\cite{kusmierz2017learning,fremaux2016neuromodulated}: the local Hebbian term $u_i x_j$ is
gated by the single scalar $g(x,u)$, which is broadcast to every synapse.
It was shown in \cite{isomura2018error} that
$\dot{W}^{(\text{EGHR})} = \dot{W}^{(\text{Oja})}\Sigma$ for Gaussian $x$, so that the two
rules share their equilibria when $\Sigma$ is non-degenerate.
We recover this identity below, but as a consequence of a structural condition rather than as
the object of the derivation, and we state precisely under which condition the stability
types also coincide (Section~\ref{sec:precond-dyn}).

\section{Frames and the cancellation lemma}
\label{sec:frames}

\subsection{Continuous frames}
We use the notion of a continuous frame \cite{ali1993continuous}, of which the familiar
discrete frames \cite{christensen2008frames} are a special case.
Let $V \subseteq \R^n$ be a subspace equipped with the standard inner product.
A random vector $\xi \in \R^n$ with $\xi \in V$ almost surely is a \emph{frame} for $V$ if
there exist constants $0 < A \leq B < \infty$ such that
\begin{alignat}{1}
    \label{Eq:frame-def}
    A|v|^2 \leq \E\big[(v, \xi)^2\big] \leq B|v|^2, \quad \forall v \in V.
\end{alignat}
A frame generalizes a basis in that it still supports expansions, at the cost of being
overcomplete.
The \emph{frame operator} is $S = \E[\xi \xi^\top]$; since $\xi \in V$ almost surely, $S$
maps $V$ into $V$ and is self-adjoint there.
Because $\E[(v,\xi)^2] = (v, Sv)$, condition~\eqref{Eq:frame-def} states exactly that $S$ is
positive definite on $V$, with $A$ and $B$ bounding its eigenvalues on $V$ from below and
above.
In particular, $S$ is invertible on $V$, and every $v \in V$ admits the frame expansion
\begin{alignat}{1}
    \label{Eq:frame-exp}
    v = \E\big[(v, S^{-1}\xi)\, \xi\big].
\end{alignat}
We stress that $S$ need not be invertible on the ambient space $\R^n$; the subspace $V$ is
where the theory lives.

\subsection{The cancellation lemma}

Eq.~\eqref{Eq:frame-exp} has exactly the structure a three-factor rule requires: the vector
$v$ is reproduced as an average of the fixed random object $\xi$, modulated by the scalar
$(v, S^{-1}\xi)$.
Read this way, $\xi$ is the local term and $(v, S^{-1}\xi)$ is the global factor.
In general, however, this factor is not useful, since it involves $S^{-1}$.
The following elementary observation identifies when it becomes simple.

\begin{lemma}[Cancellation]
\label{lem:cancel}
Let $\xi$ be a frame for $V$ with frame operator $S$.
Suppose that $v \in V$ lies in the image of $S$ restricted to $V$, that is, $v = Sm$ for some
$m \in V$.
Then
\begin{alignat}{1}
    (v, S^{-1}\xi) = (m, \xi) \qquad \text{almost surely,}
\end{alignat}
and consequently
\begin{alignat}{1}
    \label{Eq:cancel-exp}
    v = \E\big[(m, \xi)\, \xi\big].
\end{alignat}
If moreover $\E[\xi] = 0$, then $\E[(m,\xi)] = 0$.
\end{lemma}

\begin{proof}
Since $\xi \in V$ almost surely, we have $S^{-1}\xi \in V$, and since $S$ is self-adjoint on
$V$, $(Sm, S^{-1}\xi) = (m, S S^{-1}\xi) = (m, \xi)$.
Substituting this into Eq.~\eqref{Eq:frame-exp} gives Eq.~\eqref{Eq:cancel-exp}.
The last claim follows from $\E[(m,\xi)] = (m, \E[\xi]) = 0$.
\end{proof}

The content of Lemma~\ref{lem:cancel} lies not in the algebra, which is immediate, but in the
condition it isolates.
A learning rule whose flow lies in the image of a natural vector $m$ under the frame operator
automatically admits a three-factor implementation in which the global factor is the plain
inner product $(m,\xi)$---with no inverse and no distributional constants beyond those already
contained in $\xi$.
The mean subtraction that such factors typically carry is likewise not an extra ingredient:
it is inherited from $\xi$.

The remainder of this note verifies that Oja's subspace rule meets this condition and
identifies $m$ and $\xi$ in that case.

\section{Oja's subspace rule satisfies the condition}
\label{sec:oja}

\subsection{The frame}
Define
\begin{alignat}{1}
    \label{Eq:xi}
    \xi = x \otimes x - \E_x[x \otimes x] = \vct(xx^\top) - \vct(\Sigma) \in \R^{N_x^2}.
\end{alignat}
Intuitively, $\xi$ is the vectorized Hebbian term $ux^\top = Wxx^\top$ with $W$ omitted;
since $W$ enters linearly, it can be factored out.
Note that $\xi \in \vct(\Sym)$ almost surely by construction, and that $\E_x[\xi] = 0$.

\begin{proposition}
\label{prop:frame}
For Gaussian $x$ with non-degenerate covariance $\Sigma$, the random vector $\xi$ of
Eq.~\eqref{Eq:xi} is a frame for $V = \vct(\Sym)$, with frame operator
\begin{alignat}{1}
    \label{Eq:S}
    S = (\Sigma \otimes \Sigma)(I + T),
    \qquad
    S\big|_{V} = 2\,(\Sigma \otimes \Sigma)\big|_{V},
\end{alignat}
where $T$ is the commutation matrix defined by $T\,\vct(X) = \vct(X^\top)$.
The frame bounds may be taken to be $A = 2\lambda_{\min}(\Sigma)^2$ and
$B = 2\lambda_{\max}(\Sigma)^2$.
\end{proposition}

\begin{proof}
By Isserlis' theorem \cite{isserlis1918formula}, which follows from Gaussian integration by
parts, $\E_x[x_ix_jx_kx_l] = \Sigma_{ij}\Sigma_{kl} + \Sigma_{ik}\Sigma_{jl} +
\Sigma_{il}\Sigma_{jk}$.
Hence
\begin{alignat}{1}
    S &= \E_x[\xi \xi^\top]
     = \E_x[(x \otimes x)(x \otimes x)^\top] - \E_x[x \otimes x]\,\E_x[x \otimes x]^\top\\
      &= \Sigma \otimes \Sigma + (\Sigma \otimes \Sigma)T
     = (\Sigma \otimes \Sigma)(I + T).
\end{alignat}
Since $\Sigma$ is non-degenerate, it is positive definite, so $\Sigma \otimes \Sigma$ is
positive definite and in particular invertible.
Therefore $\ker S = \ker(I + T) = \vct(\Skew)$, whence $(\ker S)^\perp = \vct(\Sym) = V$.
On $V$ the commutation matrix acts as the identity, so $S|_V = 2(\Sigma \otimes \Sigma)|_V$.
Moreover, $\Sigma \otimes \Sigma$ preserves $V$, because
$(\Sigma \otimes \Sigma)\vct(X) = \vct(\Sigma X \Sigma)$ is symmetric whenever $X$ is.
The eigenvalues of $\Sigma \otimes \Sigma$ are the products
$\lambda_i(\Sigma)\lambda_j(\Sigma)$, so $S|_V$ is positive definite with spectrum contained
in $[2\lambda_{\min}(\Sigma)^2,\, 2\lambda_{\max}(\Sigma)^2]$, which yields the stated bounds
via $\E[(v,\xi)^2] = (v,Sv)$.
\end{proof}

\subsection{Preconditioning realizes the condition}

Oja's flow in the form of Eq.~\eqref{Eq:Oja-closed} is not symmetric, so it does not lie in
$V$, and Lemma~\ref{lem:cancel} cannot be applied to it directly.
Right multiplication by $\Sigma$ repairs this, and the following computation shows that it
does more: it places the flow exactly in the image of $S$, over the natural residual
$\tfrac{1}{2}(I - W^\top W)$.

\begin{proposition}
\label{prop:precond}
Let $v = \vct\big(\Sigma(I - W^\top W)\Sigma\big)$ and
$m = \vct\big(\tfrac{1}{2}(I - W^\top W)\big)$.
Then $m \in V$, $v \in V$, and
\begin{alignat}{1}
    \label{Eq:v-eq-Sm}
    v = S\,m .
\end{alignat}
\end{proposition}

\begin{proof}
Both $I - W^\top W$ and $\Sigma(I - W^\top W)\Sigma$ are symmetric, so $m, v \in V$.
By Proposition~\ref{prop:frame} and the identity $\vct(AXB) = (B^\top \otimes A)\vct(X)$,
\begin{alignat}{1}
    S\,m = 2(\Sigma \otimes \Sigma)\vct\big(\tfrac{1}{2}(I - W^\top W)\big)
         = (\Sigma \otimes \Sigma)\vct(I - W^\top W)
         = \vct\big(\Sigma(I - W^\top W)\Sigma\big) = v. \qedhere
\end{alignat}
\end{proof}

The preconditioned Oja rule is precisely $\dot{W}^{(\text{Oja})}\Sigma = W\,\vct^{-1}(v)$, so
Proposition~\ref{prop:precond} states that preconditioning is the operation that brings Oja's
rule into the form required by Lemma~\ref{lem:cancel}.
It is in this sense that the choice of $\Sigma$ is forced rather than merely convenient.
(Right multiplication by a positive definite matrix amounts to a change of metric on the
parameter space, in the spirit of natural gradient methods \cite{amari1998natural}; we return
to its dynamical effect in Section~\ref{sec:precond-dyn}.)

\subsection{The global factor}

\begin{theorem}
\label{thm:main}
Let $x$ be Gaussian with zero mean and non-degenerate covariance $\Sigma$.
Then the frame coefficient supplied by Lemma~\ref{lem:cancel} for the preconditioned Oja rule
is
\begin{alignat}{1}
    \label{Eq:g-closed}
    (m, \xi) = \Big(\tfrac{1}{2}(I - W^\top W),\; xx^\top - \Sigma\Big)_F = g(x,u),
\end{alignat}
with $g$ as defined in Eq.~\eqref{Eq:Isomura-global}, and consequently
\begin{alignat}{1}
    \label{Eq:main}
    \dot{W}^{(\text{Oja})}\Sigma = \dot{W}^{(\text{EGHR})}.
\end{alignat}
\end{theorem}

\begin{proof}
The first equality in Eq.~\eqref{Eq:g-closed} follows from the definitions of $m$ and $\xi$
together with $(\vct A, \vct B) = (A,B)_F$.
Expanding the Frobenius inner product,
\begin{alignat}{1}
    \Big(\tfrac{1}{2}(I - W^\top W),\, xx^\top - \Sigma\Big)_F
    &= \tfrac{1}{2}\tr\big((I - W^\top W)xx^\top\big)
     - \tfrac{1}{2}\tr\big((I - W^\top W)\Sigma\big)\\
    &= \tfrac{1}{2}\big(|x|^2 - |Wx|^2\big)
     - \tfrac{1}{2}\E_x\big[|x|^2 - |Wx|^2\big]\\
    &= \tfrac{1}{2}\Big(|x|^2 - |u|^2 - \E_x\big[|x|^2 - |u|^2\big]\Big) = g(x,u),
\end{alignat}
where the second line uses $\tr\big((I - W^\top W)\Sigma\big) = \E_x[x^\top(I - W^\top W)x]$.
For Eq.~\eqref{Eq:main}, apply Lemma~\ref{lem:cancel} together with
Proposition~\ref{prop:precond}:
\begin{alignat}{1}
    \vct\big(\Sigma(I - W^\top W)\Sigma\big) = v = \E_x\big[g(x,u)\,\xi\big]
    = \E_x\big[g(x,u)\,\vct(xx^\top)\big] - \E_x\big[g(x,u)\big]\vct(\Sigma),
\end{alignat}
where the last term vanishes because $\E_x[g] = \E_x[(m,\xi)] = 0$ by the final claim of
Lemma~\ref{lem:cancel}.
Undoing the vectorization and multiplying by $W$ from the left,
\begin{alignat}{1}
    \dot{W}^{(\text{Oja})}\Sigma
    = W\,\Sigma(I - W^\top W)\Sigma
    = W\,\E_x\big[g(x,u)\,xx^\top\big]
    = \E_x\big[g(x,u)\,ux^\top\big]
    = \dot{W}^{(\text{EGHR})}. \qedhere
\end{alignat}
\end{proof}

Two features of the argument deserve comment.
First, no property of $g$ was assumed: its functional form, including its mean subtraction,
was produced by Eq.~\eqref{Eq:g-closed} from $m$ and $\xi$.
Second, $g$ appears nonlinear in $x$ because of the norms $|x|^2$ and $|u|^2$, but in the
vectorized representation it is a linear functional of $\xi$, which is what allows the
expectation and the inner product to be exchanged.

\subsection{Effect of the preconditioning on the dynamics}
\label{sec:precond-dyn}

Eq.~\eqref{Eq:main} equates the two rules up to right multiplication by $\Sigma$.
The following standard fact records what this does and does not change; it makes precise the
sense in which the two rules are equivalent.

\begin{lemma}
\label{lem:precond}
Let $\Sigma$ be positive definite and let $F$ be a smooth matrix-valued map.
The flows $\dot{W} = F(W)$ and $\dot{W} = F(W)\Sigma$ have identical equilibria.
If, at an equilibrium $W_*$, the vectorized Jacobian $J = D\,\vct F(W_*)$ is symmetric, then
$J$ and the vectorized Jacobian of the second flow have the same inertia; in particular, $W_*$
is linearly stable for one flow if and only if it is linearly stable for the other.
\end{lemma}

\begin{proof}
Since $\Sigma$ is invertible, $F(W)\Sigma = 0$ if and only if $F(W) = 0$, which gives the
first claim.
For the second, $\vct(F(W)\Sigma) = (\Sigma \otimes I)\vct(F(W))$, so the Jacobian of the
second flow at $W_*$ is $J' = MJ$ with $M = \Sigma \otimes I$ positive definite.
Then $MJ = M^{1/2}\big(M^{1/2}JM^{1/2}\big)M^{-1/2}$ is similar to $M^{1/2}JM^{1/2}$, which is
symmetric and congruent to $J$; by Sylvester's law of inertia, the two have the same inertia.
\end{proof}

\begin{corollary}
Under the hypotheses of Theorem~\ref{thm:main}, EGHR-PCA and Oja's subspace rule have the same
equilibria, and the same stability type at every equilibrium at which the linearization of
Oja's rule is symmetric.
\end{corollary}

This recovers the equivalence of \cite{isomura2018error}.
We emphasize that it arises here as a by-product: the object actually derived is the global
factor, and the equivalence is what the derivation certifies.

\section{Scope}
\label{sec:scope}

The derivation uses only the frame expansion and Lemma~\ref{lem:cancel}, but three structural
features of the present setting were needed for its hypotheses to hold.
We record them to delimit the result and to indicate where extensions must do additional work.

\paragraph{Gaussian inputs.}
The computation of $S = (\Sigma \otimes \Sigma)(I + T)$ in Proposition~\ref{prop:frame} rests
on Isserlis' theorem \cite{isserlis1918formula} and is specific to Gaussian inputs.
For non-Gaussian $x$, $S$ acquires corrections involving fourth-order cumulants.
Lemma~\ref{lem:cancel} continues to apply verbatim, but the vector $m$ solving $v = Sm$ is no
longer $\tfrac{1}{2}(I - W^\top W)$, and the resulting global factor is a corrected version of
$g$.
Working out this correction explicitly is the natural next step; it would yield a three-factor
rule for non-Gaussian inputs that cannot be guessed from EGHR-PCA.

\paragraph{Bilinear Hebbian term.}
Here the Hebbian term $u_i x_j$ is bilinear in input and output, which is what allowed $W$ to
be factored out of $\xi$ in Eq.~\eqref{Eq:xi}.
More generally, Hebbian plasticity takes the form $\phi(u)\psi(x)^\top$ for nonlinear functions
$\phi$ and $\psi$.
Nonlinearity in $\phi$ makes the frame, and hence the frame operator, depend on $W$, so that
$S$ and $m$ must be determined jointly.

\paragraph{Quadratic structure of the factor.}
Both the identity $\E_x[g] = 0$ and the exchange of expectation and inner product relied on $g$
being a linear functional of $\xi$, which in turn reflects the quadratic dependence of $g$ on
$x$.
Obtaining a genuinely nonlinear third factor within this framework would require replacing the
standard inner product with a positive definite kernel, in which case $\xi$ becomes a frame in
the associated reproducing kernel Hilbert space.

\bibliographystyle{plain}
\bibliography{ref}

\end{document}